\documentclass[final]{cvpr}

\usepackage{times}
\usepackage{epsfig}
\usepackage{graphicx}
\usepackage{amsmath}
\usepackage{amssymb}
\usepackage{tabulary}
\usepackage{multirow}
\usepackage{silence}
\usepackage{caption}
\usepackage{xfrac}

\usepackage[pagebackref=true,breaklinks=true,colorlinks,bookmarks=false]{hyperref}

\newcolumntype{x}[1]{>{\centering\arraybackslash}p{#1pt}}

\newlength\savewidth\newcommand\shline{\noalign{\global\savewidth\arrayrulewidth
  \global\arrayrulewidth 1pt}\hline\noalign{\global\arrayrulewidth\savewidth}}
\newcommand{\tablestyle}[2]{\setlength{\tabcolsep}{#1}\renewcommand{\arraystretch}{#2}\centering\footnotesize}

\newcommand{\normsq}[1]{\Big\lVert#1\Big\rVert^2_2}

\def\confYear{CVPR 2021}

\begin{document}

\title{Neural Body: Implicit Neural Representations with Structured Latent Codes \\ for Novel View Synthesis of Dynamic Humans}

\author{
Sida Peng$^1$
\quad
Yuanqing Zhang$^1$
\quad
Yinghao Xu$^2$
\quad
Qianqian Wang$^3$ \\[1.5mm]
\quad
Qing Shuai$^1$
\quad
Hujun Bao$^1$
\quad
Xiaowei Zhou$^1$$^*$\\[1.5mm]
$^1$Zhejiang University
\quad
$^2$The Chinese University of Hong Kong
\quad
$^3$Cornell University
}

\twocolumn[\maketitle\vspace{-3em}
\begin{center}
\includegraphics[width=1\linewidth]{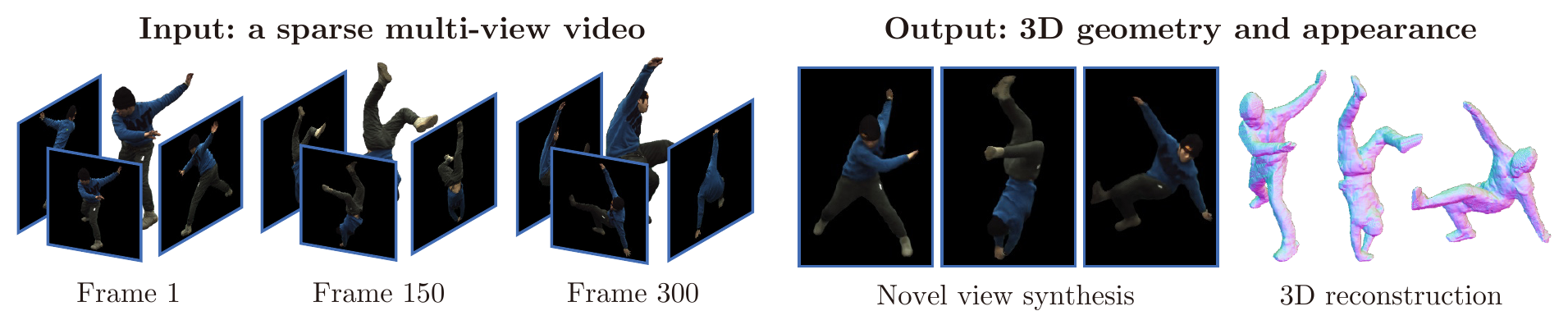}
\end{center} \vspace{-1.5em}
\captionof{figure}{\textbf{Novel view synthesis of a performer from a sparse multi-view video.} Neural Body captures the 3D geometry and appearance of the performer, which can be used for 3D reconstruction and novel view synthesis. The code and supplementary materials are available at \href{https://zju3dv.github.io/neuralbody/}{https://zju3dv.github.io/neuralbody/}.}
\label{fig:problem}
\bigbreak]

\begin{abstract}
This paper addresses the challenge of novel view synthesis for a human performer from a very sparse set of camera views.
Some recent works have shown that learning implicit neural representations of 3D scenes achieves remarkable view synthesis quality given dense input views. However, the representation learning will be ill-posed if the views are highly sparse. To solve this ill-posed problem, our key idea is to integrate observations over video frames. To this end, we propose Neural Body, a new human body representation which assumes that the learned neural representations at different frames share the same set of latent codes anchored to a deformable mesh, so that the observations across frames can be naturally integrated. The deformable mesh also provides geometric guidance for the network to learn 3D representations more efficiently.
To evaluate our approach, we create a multi-view dataset named ZJU-MoCap that captures performers with complex motions.
Experiments on ZJU-MoCap show that our approach outperforms prior works by a large margin in terms of novel view synthesis quality. We also demonstrate the capability of our approach to reconstruct a moving person from a monocular video on the People-Snapshot dataset.
\end{abstract}

\vspace{-1em}
\let\thefootnote\relax\footnotetext{The authors from Zhejiang University are affiliated with the State Key Lab of CAD\&CG. $^*$Corresponding author: Xiaowei Zhou.}
\section{Introduction}


Free-viewpoint videos of human performers have a variety of applications such as movie production, sports broadcasting, and telepresence. Previous free-viewpoint video systems either rely on a dense array of cameras for image-based novel view synthesis \cite{gortler1996lumigraph, hedman2018deep} or require depth sensors for high-quality 3D reconstruction \cite{collet2015high, dou2016fusion4d} to produce realistic rendering. The complicated hardware makes free-viewpoint video systems expensive and only applicable in constrained environments.

This work focuses on the problem of novel view synthesis for a human performer from a sparse multi-view video captured by a very limited number of cameras, as illustrated in Figure~\ref{fig:problem}. This setting significantly decreases the cost of free-viewpoint systems and makes the systems more widely applicable. However, this problem is extremely challenging. Traditional image-based rendering methods \cite{gortler1996lumigraph, debevec1996modeling} mostly require dense input views and cannot be applied here. For reconstruction-based methods \cite{schonberger2016structure, guo2019relightables}, the wide baselines between cameras make dense stereo matching intractable. Moreover, part of the human body may be invisible due to self-occlusion in sparse views. As a result, these methods tend to give noisy and incomplete reconstructions, resulting in heavy rendering artifacts.


Recent works \cite{sitzmann2019scene, niemeyer2020differentiable, mildenhall2020nerf} have investigated the potential of implicit neural representations on novel view synthesis. NeRF \cite{mildenhall2020nerf} shows that photorealistic view synthesis can be achieved by representing 3D scenes as implicit fields of density and color, which are learned from images with a differentiable renderer. However, when the input views are highly sparse, the performance of \cite{mildenhall2020nerf} degrades dramatically, as shown by our experimental results in Section~\ref{section:multi_view}. The reason is that it is ill-posed to learn the neural representations with very sparse observations. We argue that the key to solving this ill-posed problem is to aggregate all  observations over different video frames. Lombardi et al. \cite{lombardi2019neural} implement this idea by regressing the 3D representation for each frame using the same network with different latent codes as input. Since the latent codes are independently obtained for each frame, it lacks sufficient constraints to effectively fuse observations across frames. 



\begin{figure}[t]
\centering
\includegraphics[width=1\linewidth]{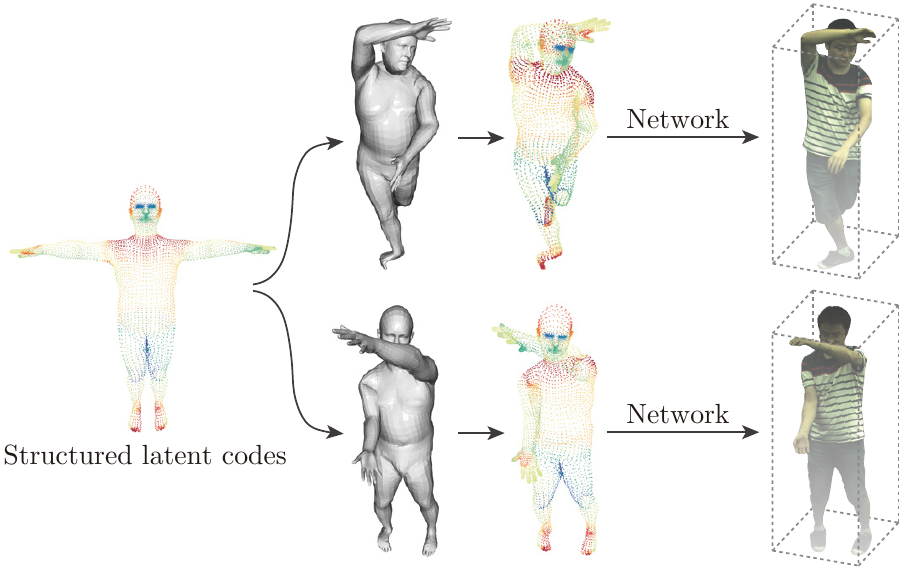}
\caption{\textbf{The basic idea of Neural Body.} Neural Body generates implicit 3D representations of a human body at different video frames from the same set of latent codes, which are anchored to the vertices of a deformable mesh. 
For each frame, we transform the spatial locations of codes based on the human pose, and use a network to regress the density and color for any 3D location based on the structured latent codes. Then, images at any viewpoints can be synthesized by the volume rendering.
}
\label{fig:basic_idea}
\vspace{-4mm}
\end{figure}

In this paper, we introduce a novel implicit neural representation for dynamic humans, named Neural Body, to solve the challenge of novel view synthesis from sparse views. The basic idea is illustrated in Figure~\ref{fig:basic_idea}. For the implicit fields at different frames, instead of learning them separately, Neural Body generates them from the same set of latent codes. Specifically, we anchor a set of latent codes to the vertices of a deformable human model (SMPL \cite{loper2015smpl} in this work), namely that their spatial locations vary with the human pose. 
To obtain the 3D representation at a frame, we first transform the code locations based on the human pose, which can be reliably estimated from sparse camera views \cite{bogo2016keep, dong2020motion, fang2021mirrored}.
Then, a network is designed to regress the density and color for any 3D point based on these latent codes.
Both the latent codes and the network are jointly learned from images of all video frames during the reconstruction process. This model is inspired by the latent variable model \cite{loehlin1987latent} in statistics, which enables us to effectively integrate observations at different frames. Another advantage of the proposed method is that the deformable model provides a geometric prior (rough surface location) to enable more efficient learning of implicit fields.

To evaluate our approach, we create a multi-view dataset called ZJU-MoCap that captures dynamic humans in complex motions. Across all captured videos, our approach exhibits state-of-the-art performances on novel view synthesis. We also demonstrate the capability of our approach to capture moving humans from monocular RGB videos on the People-Snapshot dataset \cite{alldieck2018video}. Furthermore, our approach can be used for 3D reconstruction of the performers.

In summary, this work has the following contributions:

\begin{itemize}
    \item We present a new approach capable of synthesizing photorealistic novel views of a performer in complex motions from a sparse multi-view video.
    \item We propose Neural Body, a novel implicit neural representation for a dynamic human, which enables us to effectively incorporate observations over video frames.
    \item We demonstrate significant performance improvements of our approach compared to prior work.
\end{itemize}

\begin{figure*}[t]
\centering
\includegraphics[width=1\linewidth]{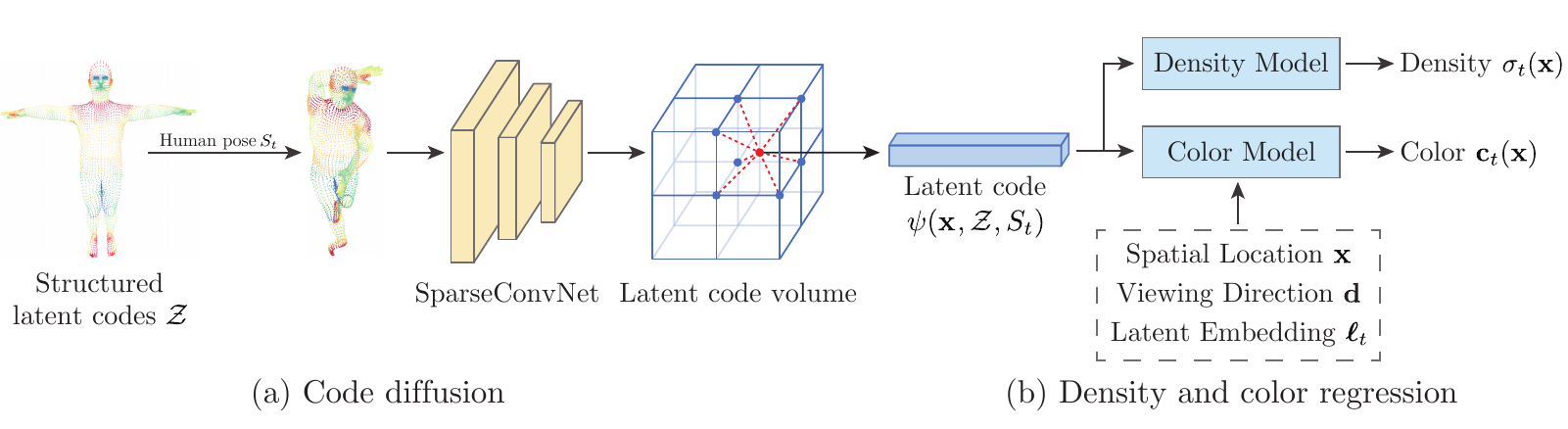}
\vspace{-2em}
\caption{\textbf{Implicit neural representation with structured latent codes.} (a) The structured latent codes are input into a SparseConvNet which outputs a latent code volume. This process diffuses the input codes defined on the surface to nearby 3D space. (b) For any 3D point, its latent code is obtained using trilinear interpolation from its neighboring vertices in the latent code volume and passed into MLP networks for density and color regression.}
\label{fig:implicit_field}
\vspace{-1em}
\end{figure*}

\section{Related work}


\paragraph{Image-based rendering.} These methods aim to synthesize novel views without recovering detailed 3D geometry. Given densely sampled images, some works \cite{gortler1996lumigraph, davis2012unstructured} apply light field interpolation to obtain novel views. Although their rendering results are impressive, the range of renderable viewpoints is limited. To extend the range, \cite{chaurasia2013depth, penner2017soft} infer depth maps from input images as proxy geometries. They utilize the depth to warp observed images into the novel view and perform image blending. However, these methods are sensitive to the quality of reconstructed proxy geometries. \cite{kalantari2016learning, hedman2018deep, choi2019extreme, thies2020ignor, kwon2020rotationally_temporally, kwon2020rotationally, wang2021ibrnet} replace hand-crafted parts of the image-based rendering pipeline with learnable counterparts to improve the robustness.

\paragraph{Human performance capture.} Most methods \cite{newcombe2015dynamicfusion, collet2015high, dou2016fusion4d, guo2019relightables} adopt the traditional modeling and rendering pipeline to synthesize novel views of humans. They rely on either depth sensors \cite{collet2015high, dou2016fusion4d, su2020robustfusion} or a dense array of cameras \cite{debevec2000acquiring, guo2019relightables} to achieve the high fidelity reconstruction. \cite{martin2018lookingood, meshry2019neural, wu2020multi} improve the rendering pipeline with neural networks, which can be trained to compensate for the geometric artifacts. To capture human models in the highly sparse multi-view setting, template-based methods \cite{carranza2003free, de2008performance, gall2009motion, stoll2010video} assume that there are pre-scanned human models. They reconstruct dynamic humans by deforming the template shapes to fit the input images. However, the deformed geometries tend to be unrealistic, and pre-scanned human shapes are unavailable in most cases. Recently, \cite{natsume2019siclope, saito2019pifu, zheng2019deephuman, saito2020pifuhd} capture the human prior from training data using networks, which enables them to recover 3D human geometry and texture from a single image. However, it is difficult for them to achieve photo-realistic view synthesis or deal with people under complex human poses that are unseen during training.

\paragraph{Neural representation-based methods.} In these works, deep neural networks are employed to learn scene representations from 2D images with differentiable renderers, such as voxels \cite{sitzmann2019deepvoxels, lombardi2019neural}, point clouds \cite{wu2020multi, aliev2020neural}, textured meshes \cite{thies2019deferred, liu2019neural, liao2020towards}, multi-plane images \cite{zhou2018stereo, flynn2019deepview}, and implicit functions \cite{sitzmann2019scene, liu2020dist, niemeyer2020differentiable, mildenhall2020nerf, liu2020neural}. As a pioneer, SRN \cite{sitzmann2019scene} proposes an implicit neural representation that maps xyz coordinates to feature vectors, and uses a differentiable ray marching algorithm to render 2D feature maps, which are then interpreted into images with a pixel generator. NeRF \cite{mildenhall2020nerf} represents scenes with implicit fields of density and color, which are well-suited for the differentiable rendering and achieve photorealistic view synthesis results. Instead of learning the scene with a single implicit function, our approach introduces a set of latent codes, which are used with a network to encode the local geometry and appearance. Furthermore, anchoring these codes to vertices of a deformable model enables us to represent a dynamic scene.

\section{Neural Body}


Given a sparse multi-view video of a performer, our task is to generate a free-viewpoint video of the performer. We denote the video as $\{\mathcal{I}_t^c | c = 1, ..., N_c, \, t = 1,...,N_t\}$, where $c$ is the camera index, $N_c$ is the number of cameras, $t$ is the frame index, and $N_t$ is the number of frames. The cameras are pre-calibrated. For each image, we apply \cite{gong2018instance} to obtain the foreground human mask and set the values of the background image pixels as zero.



The overview of the proposed model is illustrated in Figure \ref{fig:implicit_field}. Neural Body starts from a set of structured latent codes attached to the surface of a deformable human model (Section~\ref{section:structured_latent_codes}). 
The latent code at any location around the surface can be obtained with a code diffusion process (Section~\ref{section:code_diffusion}) and then decoded to density and color values by neural networks (Section~\ref{section:density_and_color}).
The image from any viewpoint can be generated by volume rendering (Section~\ref{section:volume_rendering}).
The structured latent codes and neural networks are jointly learned by minimizing the difference between the rendered images and input images (Section~\ref{section:training}).

Neural Body generates the human geometry and appearance at each frame from the same set of latent codes. From a statistical perspective, this is a type of latent variable model \cite{loehlin1987latent} that relates the observed variables at each frame to a set of latent variables. With such a latent variable model, we effectively integrate observations in the video.

\subsection{Structured latent codes}

\label{section:structured_latent_codes}

To control the spatial locations of latent codes with the human pose, we anchor these latent codes to a deformable human body model (SMPL) \cite{loper2015smpl}. SMPL is a skinned vertex-based model, which is defined as a function of shape parameters, pose parameters, and a rigid transformation relative to the SMPL coordinate system. The function outputs a posed 3D mesh with 6890 vertices. Specifically, we define a set of latent codes $\mathcal{Z} = \{\boldsymbol{z}_1, \boldsymbol{z}_2, ..., \boldsymbol{z}_{6890}\}$ on vertices of the SMPL model. For the frame $t$, SMPL parameters $S_t$ are estimated from the multi-view images $\{\mathcal{I}_t^c | c=1, ..., N_c\}$ using \cite{joo2018total}. The spatial locations of the latent codes are then transformed based on the human pose $S_t$ for the density and color regression. Figure~\ref{fig:implicit_field} shows an example. The dimension of latent code $\boldsymbol{z}$ is set to 16 in our experiments.

Similar to the local implicit representations \cite{jiang2020local, chabra2020deep, genova2020local}, the latent codes are used with a neural network to represent the local geometry and appearance of a human. Anchoring these codes to a deformable model enables us to represent a dynamic human. With the dynamic human representation, we establish a latent variable model that maps the same set of latent codes to the implicit fields of density and color at different frames, which naturally integrates observations.

\subsection{Code diffusion}

\label{section:code_diffusion}

Figure~\ref{fig:implicit_field}(a) shows the process of code diffusion. The implicit fields assign the density and color to each point in the 3D space, which requires us to query the latent codes at continuous 3D locations. This can be achieved with the trilinear interpolation. However, since the structured latent codes are relatively sparse in the 3D space, directly interpolating the latent codes leads to zero vectors at most 3D points. To solve this problem, we diffuse the latent codes defined on the surface to nearby 3D space.

\begin{table}[t]
\centering
\scalebox{0.73}{
\begin{tabular}{l|l|l}
& Layer Description & Output Dim. \\ \hline
& Input volume & D$\times$H$\times$W$\times$16 \\ \hline
1-2 & ($3 \times 3 \times 3$ conv, 16 features, stride 1) $\times$ 2 & D$\times$H$\times$W$\times$16 \\
3 & $3 \times 3 \times 3$ conv, 32 features, stride 2 & \sfrac{1}{2}D$\times$\sfrac{1}{2}H$\times$\sfrac{1}{2}W$\times$32 \\
4-5 & ($3 \times 3 \times 3$ conv, 32 features, stride 1) $\times$ 2 & \sfrac{1}{2}D$\times$\sfrac{1}{2}H$\times$\sfrac{1}{2}W$\times$32 \\
6 & $3 \times 3 \times 3$ conv, 64 features, stride 2 & \sfrac{1}{4}D$\times$\sfrac{1}{4}H$\times$\sfrac{1}{4}W$\times$64 \\
7-9 & ($3 \times 3 \times 3$ conv, 64 features, stride 1) $\times$ 3 & \sfrac{1}{4}D$\times$\sfrac{1}{4}H$\times$\sfrac{1}{4}W$\times$64 \\
10 & $3 \times 3 \times 3$ conv, 128 features, stride 2 & \sfrac{1}{8}D$\times$\sfrac{1}{8}H$\times$\sfrac{1}{8}W$\times$128 \\
11-13 & ($3 \times 3 \times 3$ conv, 128 features, stride 1) $\times$ 3 & \sfrac{1}{8}D$\times$\sfrac{1}{8}H$\times$\sfrac{1}{8}W$\times$128 \\
14 & $3 \times 3 \times 3$ conv, 128 features, stride 2 & \sfrac{1}{16}D$\times$\sfrac{1}{16}H$\times$\sfrac{1}{16}W$\times$128 \\
15-17 & ($3 \times 3 \times 3$ conv, 128 features, stride 1) $\times$ 3 & \sfrac{1}{16}D$\times$\sfrac{1}{16}H$\times$\sfrac{1}{16}W$\times$128 \\
\end{tabular}
} 
\vspace{-1mm}
\caption{\textbf{Architecture of SparseConvNet.} Each layer consists of sparse convolution, batch normalization and ReLU.}
\label{tab:architecture}
\vspace{-5mm}
\end{table}

Inspired by \cite{yan2018second, shi2020pv, peng2020convolutional}, we choose the SparseConvNet \cite{graham20183d} to efficiently process the structured latent codes, whose architecture is described in Table~\ref{tab:architecture}.
Specifically, based on the SMPL parameters, we compute the 3D bounding box of the human and divide the box into small voxels with voxel size of $5mm \times 5mm \times 5mm$. The latent code of a non-empty voxel is the mean of latent codes of SMPL vertices inside this voxel. SparseConvNet utilizes 3D sparse convolutions to process the input volume and output latent code volumes with $2\times, 4\times, 8\times, 16\times$ downsampled sizes. With the convolution and downsampling, the input codes are diffused to nearby space. Following \cite{shi2020pv}, for any point in 3D space, we interpolate the latent codes from multi-scale code volumes of network layers $5, 9, 13, 17$, and concatenate them into the final latent code. 
Since the code diffusion should not be affected by the human position and orientation in the world coordinate system, we transform the code locations to the SMPL coordinate system.

For any point $\mathbf{x}$ in 3D space, we query its latent code from the latent code volume. Specifically, the point $\mathbf{x}$ is first transformed to the SMPL coordinate system, which aligns the point and the latent code volume in 3D space. Then, the latent code is computed using the trilinear interpolation. For the SMPL parameters $S_t$, we denote the latent code at point $\mathbf{x}$ as $\psi(\mathbf{x}, \mathcal{Z}, S_t)$. The code vector is passed into MLP networks to predict the density and color for point $\mathbf{x}$.

\subsection{Density and color regression}

\label{section:density_and_color}

Figure~\ref{fig:implicit_field}(b) overviews the regression of density and color for any point in 3D space. The density and color fields are represented by MLP networks. Details of network architectures are described in the supplementary material.

\paragraph{Density model.} For the frame $t$, the volume density at point $\mathbf{x}$ is predicted as a function of only the latent code $\psi(\mathbf{x}, \mathcal{Z}, S_t)$, which is defined as:
\begin{equation}
    \sigma_t(\mathbf{x}) = M_{\sigma}(\psi(\mathbf{x}, \mathcal{Z}, S_t)),
\end{equation}
where $M_{\sigma}$ represents an MLP network with four layers.

\paragraph{Color model.} Similar to \cite{lombardi2019neural, mildenhall2020nerf}, we take both the latent code $\psi(\mathbf{x}, \mathcal{Z}, S_t)$ and the viewing direction $\mathbf{d}$ as input for the color regression. To model the location-dependent incident light, the color model also takes the spatial location $\mathbf{x}$ as input. We observe that temporally-varying factors affect the human appearance, such as secondary lighting and self-shadowing. Inspired by the auto-decoder \cite{park2019deepsdf}, we assign a latent embedding $\boldsymbol{\ell}_t$ for each video frame $t$ to encode the temporally-varying factors.

Specifically, for the frame $t$, the color at $\mathbf{x}$ is predicted as a function of the latent code $\psi(\mathbf{x}, \mathcal{Z}, S_t)$, the viewing direction $\mathbf{d}$, the spatial location $\mathbf{x}$, and the latent embedding $\boldsymbol{\ell}_t$. Following \cite{rahaman2019spectral, mildenhall2020nerf}, we apply the positional encoding to both the viewing direction $\mathbf{d}$ and the spatial location $\mathbf{x}$, which enables better learning of high frequency functions. The color model at frame $t$ is defined as:
\begin{equation}
    \mathbf{c}_t(\mathbf{x}) = M_{\mathbf{c}}(\psi(\mathbf{x}, \mathcal{Z}, S_t), \gamma_\mathbf{d}(\mathbf{d}), \gamma_\mathbf{x}(\mathbf{x}), \boldsymbol{\ell}_t),
\end{equation}
where $M_{\mathbf{c}}$ represents an MLP network with two layers, and $\gamma_\mathbf{d}$ and $\gamma_\mathbf{x}$ are positional encoding functions for viewing direction and spatial location, respectively. We set the dimension of $\boldsymbol{\ell}_t$ to 128 in experiments.

\subsection{Volume rendering}

\label{section:volume_rendering}

Given a viewpoint, we utilize the classical volume rendering techniques to render the Neural Body into a 2D image. The pixel colors are estimated via the volume rendering integral equation \cite{kajiya1984ray} that accumulates volume densities and colors along the corresponding camera ray. In practice, the integral is approximated using numerical quadrature \cite{max1995optical, mildenhall2020nerf}. Given a pixel, we first compute its camera ray $\mathbf{r}$ using the camera parameters and sample $N_k$ points $\{\mathbf{x}_k\}_{k=1}^{N_k}$ along camera ray $\mathbf{r}$ between near and far bounds. The scene bounds are estimated based on the SMPL model. Then, Neural Body predicts volume densities and colors at these points. For the video frame $t$, the rendered color $\tilde{C}_t(\mathbf{r})$ of the corresponding pixel is given by:
\begin{gather}
    \tilde{C}_t(\mathbf{r}) = \sum_{k=1}^{N_k} T_k (1 - \exp(-\sigma_t(\mathbf{x}_k) \delta_k)) \mathbf{c}_t(\mathbf{x}_k), \\
    \text{where} \quad T_k = \exp(-\sum_{j=1}^{k-1} \sigma_t(\mathbf{x}_j) \delta_j),
\end{gather}
where $\delta_k = || \mathbf{x}_{k + 1} - \mathbf{x}_{k} ||_2$ is the distance between adjacent sampled points. We set $N_k$ as 64 in all experiments. With volume rendering, our model is optimized by comparing the rendered and observed images.

\subsection{Training}

\label{section:training}

Through the volume rendering techniques, we optimize the Neural Body to minimize the rendering error of observed images $\{\mathcal{I}_t^c | c = 1, ..., N_c, \, t = 1,...,N_t\}$:
\begin{equation}
    \underset{\{\boldsymbol{\ell}_t\}_{t=1}^{N_t}, \mathcal{Z}, \Theta \,}{\text{minimize}} \sum_{t=1}^{N_t} \sum_{c=1}^{N_c} L(\mathcal{I}_t^c, P^c; \boldsymbol{\ell}_t, \mathcal{Z}, \Theta),
    \label{eq:optimization}
\end{equation}
where $\Theta$ means the network parameters, $P^c$ is the camera parameters, and $L$ is the total squared error that measures the difference between the rendered and observed images. The corresponding loss function is defined as:
\begin{equation}
    L = \sum\limits_{\mathbf{r} \in \mathcal{R}} \normsq{\tilde{C}(\mathbf{r}) - C(\mathbf{r})},
\end{equation}
where $\mathcal{R}$ is the set of camera rays passing through image pixels, and $C(\mathbf{r})$ means the ground-truth pixel color. In contrast to frame-wise reconstruction methods \cite{schonberger2016structure, mildenhall2020nerf}, our method optimizes the model using all images in the video and has more information to recover the 3D structures.

We adopt the Adam optimizer \cite{kingma2014adam} for training the Neural Body. The learning rate starts from $5e^{-4}$ and decays exponentially to $5e^{-5}$ along the optimization. We conduct the training on four 2080 Ti GPUs. The training on a four-view video of 300 frames typically takes around 200k iterations to converge (about 14 hours).

\subsection{Applications}

The trained Neural Body can be used for novel view synthesis and 3D reconstruction of the performer. The view synthesis is achieved through the volume rendering. Novel view synthesis on dynamic humans results in free-viewpoint videos, which give the viewers the freedom to watch human performers from arbitrary viewpoints. Our experimental results show that the generated videos exhibit high inter-frame and inter-view consistency, which are presented in the supplementary material. For 3D reconstruction, we first discretize the scene with a voxel size of $5mm \times 5mm \times 5mm$. Then, we evaluate the volume densities for all voxels and extract the human mesh with the Marching Cubes algorithm \cite{lorensen1987marching}.

\section{Experiments}

\subsection{Results on the ZJU-MoCap dataset}

\label{section:multi_view}

We create a multi-view dataset called ZJU-Mocap for evaluating our approach. This dataset captures 9 dynamic human videos using a multi-camera system that has 21 synchronized cameras. We select four uniformly distributed cameras for training and use the remaining cameras for test. All sequences have a length between 60 to 300 frames. The humans perform complex motions, including twirling, Taichi, arm swings, warmup, punching, and kicking.

\paragraph{Metrics.} For novel view synthesis, we follow \cite{mildenhall2020nerf} to evaluate our method using two standard metrics: peak signal-to-noise ratio (PSNR) and structural similarity index (SSIM). For 3D reconstruction, we only provide qualitative results, as there is no ground-truth human geometry.

\begin{table}
\begin{center}
\scalebox{0.73}{
\tablestyle{4pt}{1.05}
\begin{tabular}{c|x{28}x{28}x{34}x{20}|x{28}x{28}x{34}x{20}}
& \multicolumn{4}{c|}{PSNR} & \multicolumn{4}{c}{SSIM} \\[.1em]
\shline
& NV \cite{lombardi2019neural} & NT \cite{thies2019deferred} & NHR \cite{wu2020multi} & OURS & NV \cite{lombardi2019neural} & NT \cite{thies2019deferred} & NHR \cite{wu2020multi} & OURS \\
\hline
Twirl  & 22.09 & 25.78 & 26.68 & \textbf{30.56} & 0.831 & 0.929 & 0.935 & \textbf{0.971} \\
Taichi & 18.57 & 19.44 & 19.81 & \textbf{27.24} & 0.824 & 0.869 & 0.874 & \textbf{0.962} \\
Swing1 & 22.88 & 24.96 & 24.73 & \textbf{29.44} & 0.726 & 0.905 & 0.902 & \textbf{0.946} \\
Swing2 & 22.08 & 24.84 & 25.01 & \textbf{28.44} & 0.843 & 0.903 & 0.906 & \textbf{0.940} \\
Swing3 & 21.29 & 23.50 & 23.47 & \textbf{27.58} & 0.842 & 0.896 & 0.894 & \textbf{0.939} \\
Warmup & 21.15 & 23.74 & 23.79 & \textbf{27.64} & 0.842 & 0.917 & 0.918 & \textbf{0.951} \\
Punch1 & 23.21 & 24.93 & 25.02 & \textbf{28.60} & 0.820 & 0.877 & 0.879 & \textbf{0.931} \\
Punch2 & 20.74 & 22.44 & 22.88 & \textbf{25.79} & 0.838 & 0.888 & 0.891 & \textbf{0.928} \\
Kick   & 22.49 & 24.33 & 23.72 & \textbf{27.59} & 0.825 & 0.881 & 0.873 & \textbf{0.926} \\
\hline                                                
average& 21.39 & 23.77 & 23.90 & \textbf{28.10} & 0.821 & 0.896 & 0.897 & \textbf{0.944} \\
\end{tabular}
}
\end{center}
\vspace{-1em}
\caption{\textbf{Results on the ZJU-MoCap dataset in terms of PSNR and SSIM (higher is better).} ``NV" means Neural Volumes, and ``NT" means Neural Textures.}
\label{table:multi_view}
\end{table}


\begin{figure*}[t]
\centering
\includegraphics[width=1\linewidth]{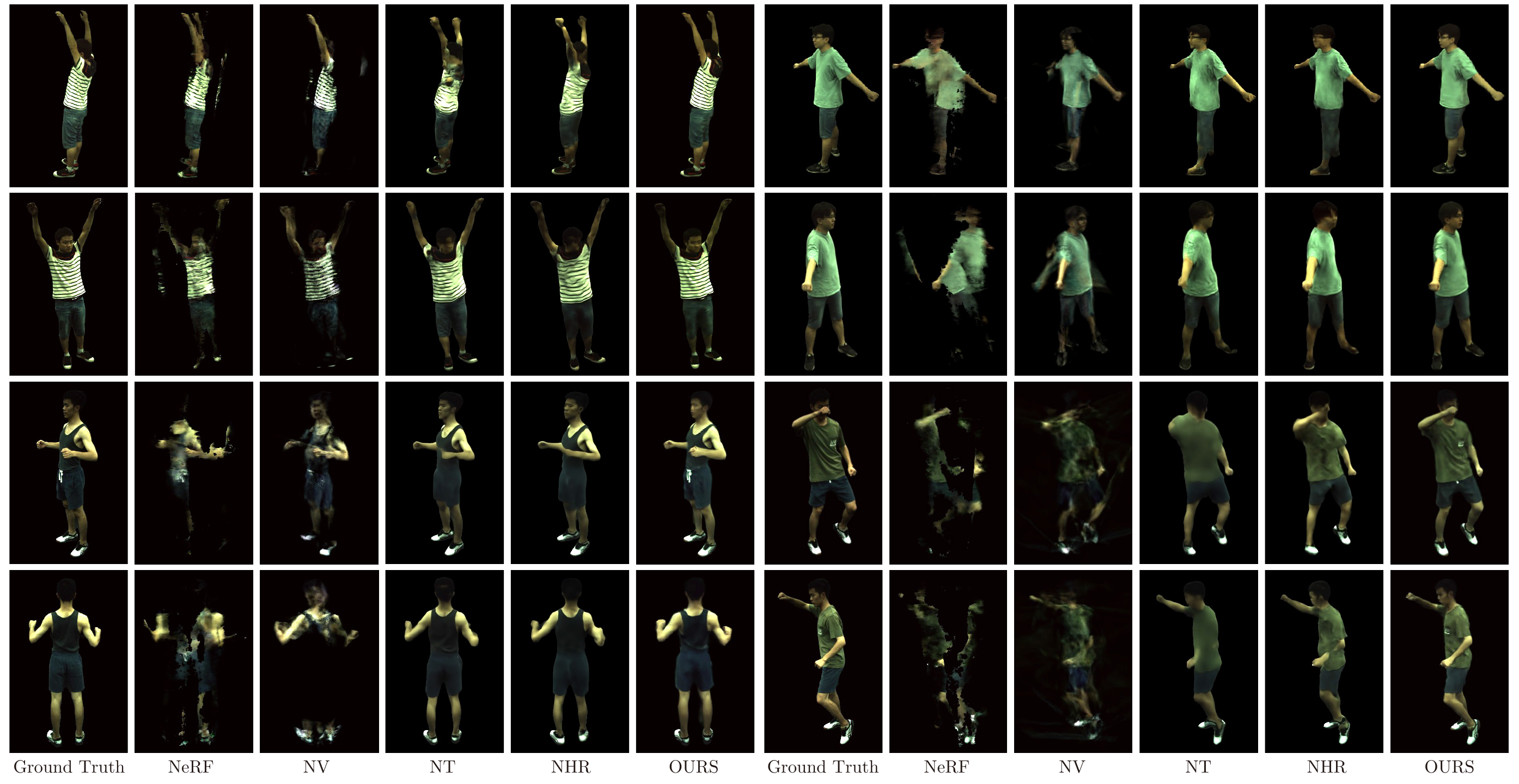}
\caption{\textbf{Novel view synthesis on the ZJU-MoCap dataset.} ``NV" means Neural Volumes \cite{lombardi2019neural}, and ``NT" means Neural Textures \cite{thies2019deferred}. The input video is captured by four cameras. We select two novel views for qualitative comparison. Our method significantly outperforms \cite{lombardi2019neural, mildenhall2020nerf}. Furthermore, compared with image-to-image translation methods \cite{thies2019deferred, wu2020multi}, we can produce temporally consistent free-viewpoint videos, which are presented in the supplementary material.}
\label{fig:multi_view_novel_view}
\vspace{-2mm}
\end{figure*}

\paragraph{Performance on novel view synthesis.} We compare our method with state-of-the-art view synthesis methods \cite{lombardi2019neural, thies2019deferred, wu2020multi} that handle dynamic scenes. All methods train a separate network for each scene. 1) Neural Volumes \cite{lombardi2019neural} encodes multi-view images at each frame into a latent vector and decodes it into a discretized RGB$\alpha$ voxel grid. 2) Neural Textures \cite{thies2019deferred} proposes latent texture maps to render a coarse mesh into 2D images. Since \cite{thies2019deferred} is not open-sourced, we reimplement it and take the SMPL mesh as the input mesh. 3) NHR \cite{wu2020multi} uses networks to render input point clouds to images. Here we take SMPL vertices as input point clouds.

Table~\ref{table:multi_view} shows the comparison of our method with \cite{lombardi2019neural, thies2019deferred, wu2020multi} in terms of the PSNR metric and the SSIM metric, respectively. For both metrics, our model achieves the best performances among all methods. In particular, our method outperforms previous works by a margin of at least 4.20 in terms of PSNR and 0.047 in terms of SSIM.

In contrast to learning the 3D representations from individual latent vectors \cite{lombardi2019neural}, Neural Body generates implicit fields at different frames from the same set of latent codes. The results indicate that our method better integrates observations of the target performer across video frames.

Figure~\ref{fig:multi_view_novel_view} shows the qualitative results of our method and other methods \cite{lombardi2019neural, thies2019deferred, wu2020multi, mildenhall2020nerf}. Here NeRF \cite{mildenhall2020nerf} trains a separate network for each video frame. The rendering results of \cite{mildenhall2020nerf, lombardi2019neural} indicate that they don't accurately capture the 3D human geometry and appearance. The results of NeRF \cite{mildenhall2020nerf} don't appear reasonable shapes, which show that NeRF fails to learn correct 3D human representations. Neural Volumes \cite{lombardi2019neural} gives blurry results. As image-to-image translation methods, \cite{thies2019deferred, wu2020multi} have difficulty in controlling the rendering viewpoints. In contrast, our method gives photorealistic novel views. Furthermore, our method can generate inter-frame and inter-view consistent free-viewpoint videos, which are presented in the supplementary material.

\begin{figure*}[t]
\centering
\includegraphics[width=1\linewidth]{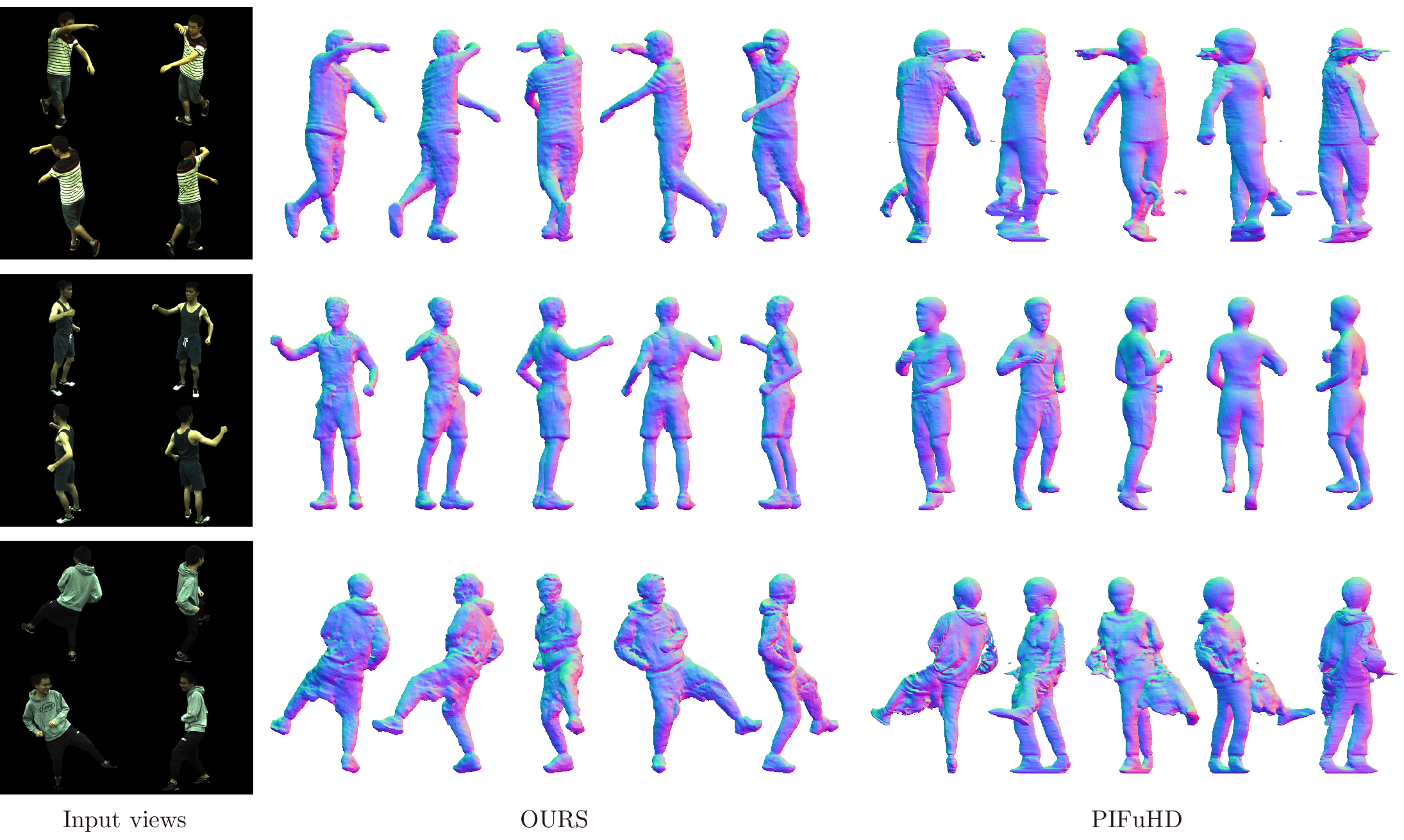}
\vspace{-1.5em}
\caption{\textbf{3D reconstruction on the ZJU-MoCap dataset.} Neural Body achieves high-quality reconstruction. Our method is able to recover the clothing, such as the hoodie of the third person. PIFuHD \cite{saito2020pifuhd} does not generalize well on the dataset.}
\label{fig:multi_view_reconstruction}
\vspace{-2mm}
\end{figure*}

\paragraph{Performance on 3D reconstruction.} We test state-of-the-art multi-view methods COLMAP \cite{schonberger2016structure, schonberger2016pixelwise} and DVR \cite{niemeyer2020differentiable} on the ZJU-MoCap dataset. COLMAP \cite{schonberger2016structure, schonberger2016pixelwise} is a well-developed multi-view stereo algorithm, and DVR \cite{niemeyer2020differentiable} learns occupancy fields \cite{mescheder2019occupancy} with a differentiable renderer. We find that they fail to recover reasonable 3D human shapes from only four input views.

For comparison, we choose a learning-based approach, PIFuHD \cite{saito2020pifuhd}, as the baseline method. PIFuHD trains a single-view reconstruction network on 450 high-resolution photogrammetry scans. We use its released code and pre-trained model for inference. The first view is taken as the input of PIFuHD. To improve its performance, we remove the background of the input image. \cite{huang2018deep, saito2019pifu} propose multi-view reconstruction networks, but they don't release the pre-trained model, so we don't compare with them.

Figure~\ref{fig:multi_view_reconstruction} presents the qualitative comparison between our method and PIFuHD. Neural Body generates accurate geometries for humans in complex motions. Since our method learns the 3D human representations from multi-view images, the 3D human poses of reconstructed human models are highly consistent with the observations. The reconstruction results of PIFuHD indicate that it doesn't generalize well on our data. For persons with complex human poses, PIFuHD fails to recover correct human shapes. Moreover, its reconstructed models are not consistent with the human poses observed from multi-view images.

\begin{figure*}[t]
\centering
\includegraphics[width=1\linewidth]{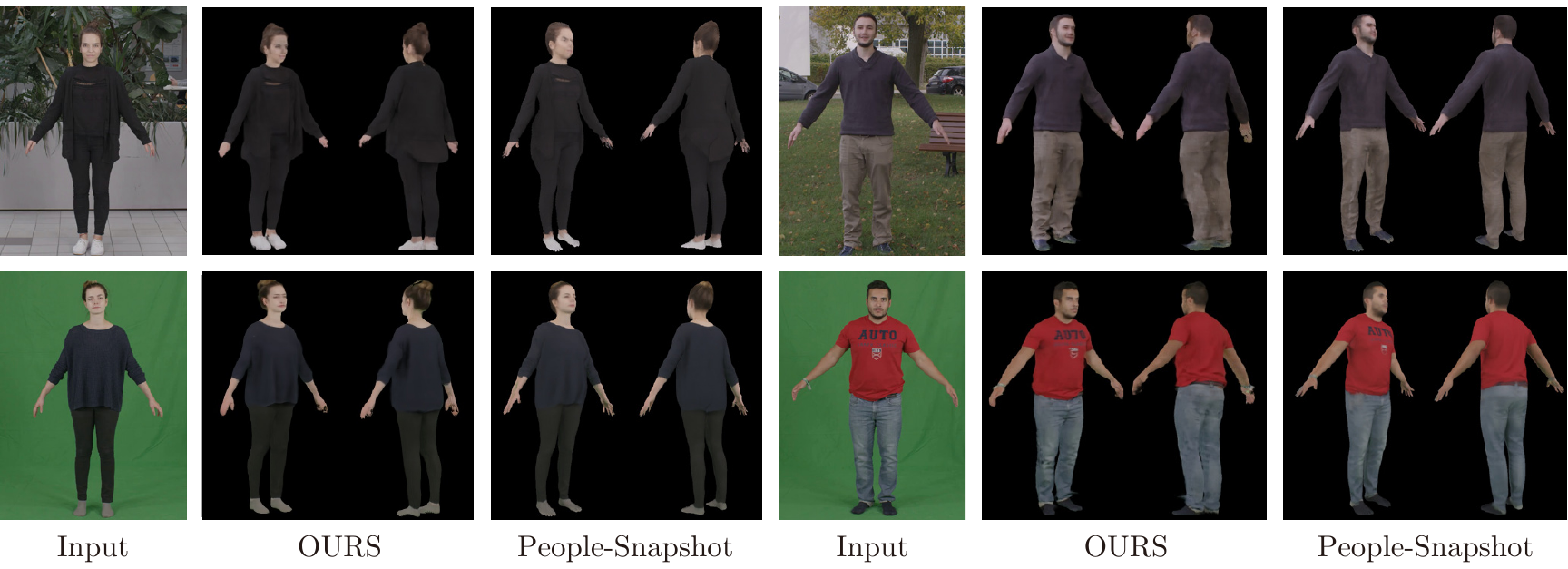}
\vspace{-1.7em}
\caption{\textbf{Novel view synthesis on monocular videos.} Our method renders more appearance details than People-Snapshot \cite{alldieck2018video}, such as the blouse of the first person and the pants of the second person. Zoom in for details.}
\vspace{-1em}
\label{fig:people_snapshot_novel_view}
\end{figure*}

\begin{figure*}[t]
\centering
\includegraphics[width=1\linewidth]{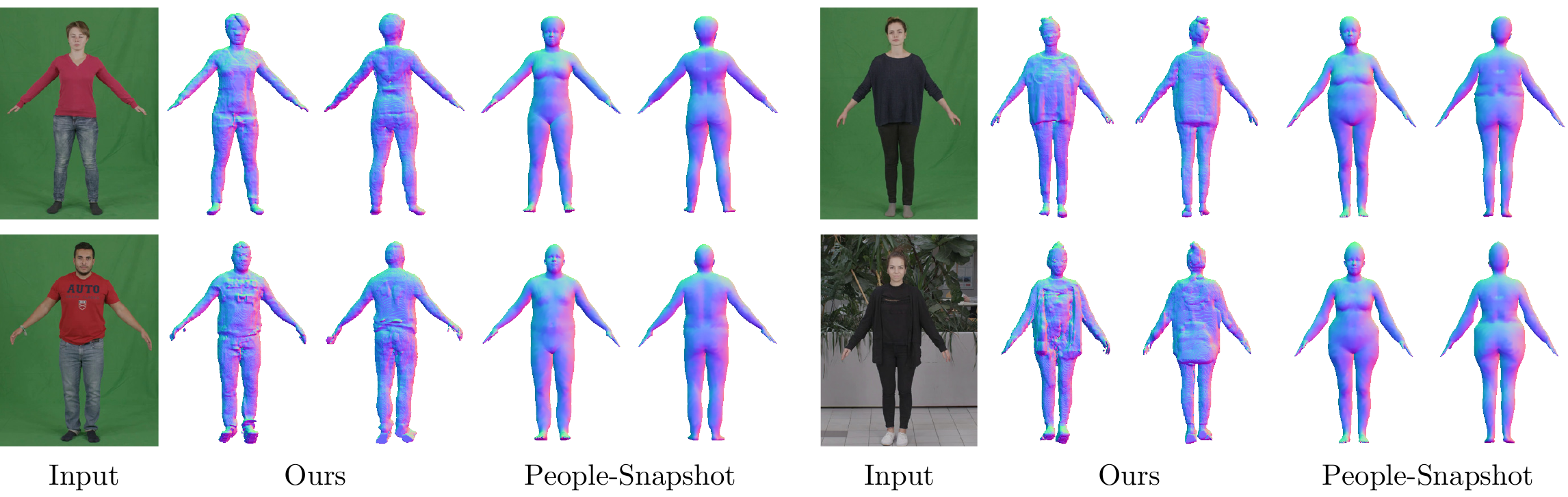}
\vspace{-1.7em}
\caption{\textbf{3D reconstruction on monocular videos.} Compared with the approach in People-Snapshot \cite{alldieck2018video}, Neural Body generates more detailed geometries and can handle persons wearing loose clothing.}
\label{fig:people_snapshot_reconstruction}
\end{figure*}

\subsection{Results on monocular videos}

We demonstrate that our approach is able to reconstruct dynamic humans from monocular videos on the People-Snapshot dataset \cite{alldieck2018video}. This dataset captures performers that rotate while holding an A-pose. Since the poses of moving humans are not complex, the SMPL parameters can be accurately estimated from the monocular videos. We compare Neural Body with the approach proposed in \cite{alldieck2018video}, which deforms vertices of the SMPL model to fit the 2D human silhouettes over the video sequence. Following \cite{alldieck2018video}, we report the qualitative results on the People-Snapshot dataset.

\paragraph{Performance on novel view synthesis.} Figure~\ref{fig:people_snapshot_novel_view} shows the qualitative comparison on novel view synthesis. Our method renders more appearance details than \cite{alldieck2018video}, especially for the performers wearing the loose clothing. For example, Neural Body accurately renders the blouse for the first person, while the blouse rendered by \cite{alldieck2018video} attaches closely to the human body. Some of scenes are captured in the outdoor environment, which exhibit strong illumination variations. The photorealistic rendering results indicate that Neural Body can handle complex lighting conditions.

\paragraph{Performance on 3D reconstruction.} The qualitative results of our method and \cite{alldieck2018video} are presented in Figure~\ref{fig:people_snapshot_reconstruction}. Neural Body recovers more geometric details than \cite{alldieck2018video}. For example, the hair shapes are highly consistent with the RGB observations. The results of the last column indicate that our method can handle persons wearing loose clothing, while \cite{alldieck2018video} does not recover correct shapes for such data.

\begin{table}
\begin{center}
\tablestyle{4pt}{1.05}
\begin{tabular}{x{30}|x{35}|x{35}|x{35}|x{35}}
& 1 view & 2 views & 4 views & 6 views \\[.1em]
\shline
PSNR & 25.08 & 25.49 & 30.54 & \textbf{32.73} \\
SSIM & 0.912 & 0.928 & 0.971 & \textbf{0.979} \\
\end{tabular}
\vspace{-1.5em}
\end{center}
\caption{\textbf{Results of models trained with different numbers of camera views} on the video ``Twirl" of the ZJU-MoCap dataset. We select six camera views for ablation studies and use the remaining views for test.}
\vspace{-0.5em}
\label{table:ablation}
\end{table}

\subsection{Ablation studies on the ZJU-Mocap dataset}

We conduct ablation studies on the video ``Twirl". We first analyze the effects of per-frame latent embedding. Then we explore the performances of our models trained with different numbers of video frames and input views.

\textbf{Impact of per-frame latent embedding.} We train a model without latent embeddings $\{\boldsymbol{\ell}_t\}_{t=1}^{N_t}$ that are proposed in Section~\ref{section:density_and_color}, which gives $30.03$ PSNR, lower than $30.56$ PSNR of the complete model. This comparison indicates that the latent embeddings yield $0.53$ PSNR improvement.

\textbf{Impact of the number of camera views.} Table~\ref{table:ablation} compares our models trained with different numbers of camera views. The results show that the number of training views improves the performance on novel view synthesis. Neural Body trained on single view still outperforms \cite{lombardi2019neural} trained on four views, which gives $23.12$ PSNR and $0.875$ SSIM on test views of the ablation study.

\textbf{Impact of the video length.} We train our model with 1, 60, 300, 600, and 1200 frames, respectively. The results are evaluated on the first frame of the video ``Twirl". Table~\ref{table:num_frames} shows the quantitative results, which indicate that training on the video improves the view synthesis performance, but training on too many frames may decrease the performance as the network has difficulty in fitting very long videos.

\begin{table}
\begin{center}
\tablestyle{4pt}{1.05}
\begin{tabular}{x{30}|x{32}|x{32}|x{32}|x{32}|x{32}}
Frames & 1 & 60 & 300 & 600 & 1200 \\[.1em]
\shline
PSNR & 25.64 & 30.14 & \textbf{30.66} & 30.59 & 29.97 \\
SSIM & 0.940 & 0.970 & \textbf{0.971} & 0.970 & 0.970
\end{tabular}
\vspace{-1.5em}
\end{center}
\caption{\textbf{Results of models trained with different numbers of training frames.} We train models on 1, 60, 300, 600, and 1200 frames and test on the first frame of ``Twirl".}
\label{table:num_frames}
\end{table}

\section{Conclusion}

We introduced a novel implicit neural representation, named Neural Body, for novel view synthesis of dynamic humans from sparse multi-view videos. Neural Body defines a set of latent codes, which encode local geometry and appearance with a neural network. We anchored these latent codes to vertices of a deformable human model to represent a dynamic human. This enables us to establish a latent variable model that generates implicit fields at different video frames from the same set of latent codes, which effectively incorporates observations of the performer across video frames. We learned Neural Body over the video with volume rendering. To evaluate our approach, we created a multi-view dataset called ZJU-MoCap that captures dynamic humans in complex motions. We demonstrated superior view synthesis quality compared to prior work on the newly collected dataset and the People-Snapshot dataset.

\vspace{1em}
\noindent\textbf{Acknowledgements:} 
The authors from Zhejiang University would like to acknowledge the support from the National Key Research and Development Program of China (No. 2020AAA0108901) and NSFC (No. 61806176).

{\small
\bibliographystyle{ieee_fullname}
\bibliography{egbib}
}

\end{document}